\documentclass[preprint,12pt]{elsarticle}

\usepackage{amssymb}
\usepackage[dvipsnames]{xcolor}
\usepackage{pgfplots}
\pgfplotscreateplotcyclelist{fcycle}{%  %<------
    {blue}, %<------
    {orange},
    {green},%<------
    {red},%<------
    {violet},
    {teal},
}
\usepackage[utf8]{inputenc}
\usepackage[T1]{fontenc}
\usepackage[english]{babel}
\usepackage{indentfirst}
\usepackage{enumitem}
\usepackage{amssymb}
\usepackage{amsmath}
\usepackage{amsfonts}
\usepackage{multicol}
\usepackage{mathrsfs}
\usepackage{stmaryrd}
\usepackage{lscape}
\usepackage{colortbl}
\usepackage{fancyhdr}
\usepackage{tabularx,environ}
\usepackage{appendix}
\usepackage{array}
\usepackage{multirow}
\usepackage{colortbl}
\usepackage{epsfig}
\usepackage{indentfirst}
\usepackage{amsmath}
\usepackage{amsfonts}
\usepackage{amssymb}
\usepackage{calrsfs}
\usepackage{caption}
\usepackage{subcaption}
\usepackage{color}
\usepackage{amsmath}
\usepackage{amsthm}
\usepackage{amssymb}
\usepackage{graphicx}
\usepackage{systeme}
\usepackage{lmodern}
\usepackage{mathtools}
\pgfplotsset{compat=1.18} 
\usepackage[a4paper,left=3.2cm,right=3.2cm,top=3cm,bottom=3cm]{geometry}
\makeatletter
\DeclareRobustCommand{\iscircle}{\mathord{\mathpalette\is@circle\relax}}
\newcommand\is@circle[2]{%
  \begingroup
  \sbox\z@{\raisebox{\depth}{$\m@th#1\bigcirc$}}%
  \sbox\tw@{$#1\square$}%
  \resizebox{!}{\ht\tw@}{\usebox{\z@}}%
  \endgroup
}
\makeatother
\journal{Arxiv}

\begin{document}

\begin{frontmatter}



\title{A Virtual-Force Based Swarm Algorithm for Balanced Circular Bin Packing Problems}


\author[inst1,inst2]{Juliette Gamot}

\affiliation[inst1]{organization={ONERA/DTIS},%Department and Organization
            addressline={Université Paris-Saclay}, 
            city={Palaiseau},
            country={France}}

\author[inst1]{Mathieu Balesdent}
\author[inst1]{Romain Wuilbercq}
\author[inst1]{Arnault Tremolet}

\author[inst2]{Nouredine Melab}
\author[inst2]{El-Ghazali Talbi}

\affiliation[inst2]{organization={INRIA},%Department and Organization
            addressline={Université de Lille}, 
            city={Villeneuve d'Ascq}, 
            country={France}}

\begin{abstract}
%Balanced circular bin packing problems involves positioning a given number of weighted circles in order to minimize the radius of the circular container while satisfying equilibrium constraints. This paper describes a swarm algorithm based on a virtual-force system in order to solve the aforementioned problem. In the proposed approach, a system of forces is applied to each component allowing to take into account the constraints and minimizing the objective function by applying the Newton's second law of motion.
%The proposed strategy is illustrated on various instances of balanced circular bin packing problems with 10 to 300 circles. The performance of the algorithm is analyzed and the final best obtained layouts are compared with existing results from the literature.

Balanced circular bin packing problems consist in positioning a given number of weighted circles in order to minimize the radius of a circular container while satisfying equilibrium constraints. These problems are NP-hard, highly constrained and dimensional. This paper describes a swarm algorithm based on a virtual-force system in order to solve balanced circular bin packing problems. In the proposed approach, a system of forces is applied to each component allowing to take into account the constraints and minimizing the objective function using the fundamental principle of dynamics. The proposed algorithm is experimented and validated on benchmarks of various balanced circular bin packing problems with up to 300 circles. The reported results allow to assess the effectiveness of the proposed approach compared to existing results from the literature.

\end{abstract}




\end{frontmatter}


\section{Introduction}
\subsection{General context}
Packing problems consist in positioning a set of objects with various sizes into one or several container(s), while satisfying constraints such as capacity or weight balancing. Various configurations of packing problems are defined depending on the shape of the items: circles \cite{Grosso,Yuan,Stoyan}, rectangles \cite{Bengtsson,Chen,Bouzid} or polygons \cite{Daoden,Luo,Guerriero},  or depending on the geometry of the container: circles \cite{Yuan,Castillo}, rectangles \cite{He-Tole,Huang}, squares or polygons \cite{Galiev,Lopez,Birgin},  as well as depending on the number of bins: single bin \cite{He,Liu1} or multiple bins \cite{Yuan,He-Tole}.

Packing problems have practical applications in various domains, such as complex systems design, transportation as well as facility dispersion and communication networks. In complex systems design (\textit{e.g.} automotive, aerospace vehicles), a set of components has to be packed within a container while minimizing for instance the used space and satisfying several constraints including equilibrium ones \cite{Sun, Li2, Gamot1}. In transportation, different items need to be packed into vehicles of different sizes while for example minimizing the empty space within the vehicles, minimizing the number of used vehicles, balancing each vehicle, \textit{etc.} \cite{Mongeau,Gajda,Caprace}. In facility dispersion and communication networks, the aim is to disperse some facilities in a given area in order for example to maximize a communication or network coverage, minimize delivery distances from warehouses to shipping addresses, maximize the energy production of a wind farm, \textit{etc}. \cite{Adickes, Boonmee,Herbert}.

Among all the possible layout problems, in this paper, the balanced circular bin packing problem is considered. It consists in packing a set of circular items with different sizes into a circular container, while ensuring that the weight of the items is balanced. The objective is to minimize the size of the container (\textit{i.e.} maximize the occupation rate of the bin) used to pack all items, while satisfying the balancing constraint \cite{Liu1,Xiao,Xu}.

\subsection{Related works}

Bin packing problems are NP-hard problems that are widely documented in the literature. 
From a mathematical point of view,  the rectangular bin packing problem has been addressed by Costa \textit{et al.} \cite{Costa}  by means of an adapted branch-and-bound algorithm.  Stoyan \textit{et al.} \cite{Stoyan} proposed a mathematical model with an insertion strategy to pack equal circles into circular containers with prohibited zones.

Various exact or approximate methods have been explored to solve packing problems. Akeb \textit{et al.} \cite{Akeb} packed unequal circles into a circular container using an adaptive beam search algorithm that combines beam search with the notion of local position distance and dichotomous search. Galiev \textit{et al.} \cite{Galiev} proposed a heuristic algorithm based on linear models for packing equal circles into containers of various shapes. In \cite{Guerriero}, the authors developed a two-level heuristic called hierarchical hyper-heuristic in order to position irregular polygons into squared or rectangular containers.
In \cite{Huang}, a quasi-physical (\textit{i.e.} with both physical and mathematical roots) global optimization method is described. A local search strategy based on an energetic and physical modeling of the circles has been coupled with a basin-hopping procedure to escape local optimum traps. This quasi-physical strategy has been extended in \cite{He-Ye} using a modified Broyden–Fletcher–Goldfarb–Shanno algorithm and a new basin-hopping strategy to solve unequal circular bin packing problems. Zeng \textit{et al.} \cite{Zeng} developed an iterated tabu search and variable neighborhood descent in order to solve the same problem.

Packing problems have also been solved using population-based or hybrid approaches. Luo \textit{et al.} \cite{Luo} applied a genetic and a grey wolf optimization algorithm coupled to a heuristic strategy to tackle packing problems with irregular polygonal items and containers. In \cite{Xu}, the authors applied a new particle swarm optimization algorithm coupled with local search to optimize the layout of rectangular items into circular containers.

More recently Machine Learning methods have been applied to packing problems. Bello \textit{et al.} \cite{Bello} tackled combinatorial optimization problems including the knapsack problem with a reinforcement learning strategy.
In \cite{Kundu}, \cite{Zhao} and \cite{Hu}, constrained deep reinforcement learning is applied to 2-dimensional and 3-dimensional bin packing problems.\\

This paper focuses on a particular formulation of bin packing problems: the balanced circular bin packing problem, also referred as packing problem with equilibrium constraints and weighted or load-balanced packing problems. 
Table \ref{tab:biblio} reviews the main contributions regarding packing problems with balancing constraints as well as their main characteristics.\\

\begin{table}[h!]
\centering
  \begin{tabularx}{1.0\textwidth}{ 
  | >{\centering\arraybackslash}m{0.8cm} 
  | >{\centering\arraybackslash}m{1.2cm} 
  | >{\centering\arraybackslash}m{2.1cm} 
  | >{\centering\arraybackslash}m{2cm} 
  | >{\centering\arraybackslash}m{6.3cm} |}
    \hline
    Ref. & Items  & Container(s) & Objective &  Method \\
    \hline
    \hline
    \cite{He} & $ \iscircle$ & $\iscircle$ & V & Quasi-Physical model and Tabu Search (TS)\\
    \hline
    \cite{Liu1} & $ \iscircle$ & $\iscircle$ & V & Energy Landscape Paving (ELP), local search, heuristic\\
    \hline
    \cite{Xiao} & $ \iscircle$ & $\iscircle$ & V & Simulated Annealing (SA) and Particle Swarm Optimization (PSO) \\
    \hline
    \cite{Xu} & $ \square$ & $\iscircle$ & V & PSO \\
    
    \hline
    \cite{Tang} & $\iscircle $ & $\iscircle$ & V & Genetic Algorithm (GA) \\
    \hline
    \cite{Teng} & $\iscircle , \square$ & $\iscircle$ & I & Heuristic and mathematical programming \\
    \hline

    \cite{Zhang} & $ \iscircle, \square$ & $\iscircle$ & I & Neural Network, GA, PSO, Quasi-principal component analysis\\
    \hline
    
    \cite{Liu2} & $ \iscircle$ & $\iscircle$ & V & Hybrid algorithm: SA + heuristic neighborhood search + adaptive gradient method\\
    \hline
    \cite{Liu3} & $ \iscircle$ & $\iscircle$ & V & Hybrid algorithm: Heuristic + Tabu Search (TS)\\
    \hline
    \cite{Li} & $ \iscircle$ & $\square$ & V & Knowledge-based heuristic PSO\\
    \hline

    \cite{Wang} & $ \iscircle$ & $\iscircle$ & V & PSO, Ant Colony Optimization, Artificial Bee Colony\\
    \hline
    \cite{Moon} & $ \square$ & $\square$ & V & Hybrid algorithm: Greedy strategy + GA\\
    \hline
    \cite{Liu4} & $ \iscircle$ & $\iscircle$ & V & ELP + local search + conformation update strategy\\
    \hline
    \cite{Liu5} & $ \iscircle$ & $\iscircle,\square$ & V & Heuristic quasi-physical algorithm\\
    \hline
    \cite{Liu6} & $ \square$ & $\iscircle$ & V & ELP and local search \\
    \hline
    \cite{Wang2} & $ \iscircle$ & $\iscircle$ & V & Stimulus–response-based allocation (SRA) method \\
    \hline
    \cite{Erbayrak} & $ \square$ & $\square$ & B, 3D & Multi-objective mixed integer programming model\\
    \hline
    \cite{Romanova1} & $ \diamondsuit $ & $\iscircle$ & V, 3D & Nonlinear Programming model\\
    \hline
    \cite{Romanova2} & $ \iscircle$ & $\iscircle$ & V & Sequential algorithm based on Shor's r-algorithm\\

    \hline
  \end{tabularx}
  \caption{Review of the main contributions for balanced packing problems. Items or containers: Circles ($\iscircle)$, Squares/Rectangles($\square$), Polygons($\diamondsuit$). Objective: Minimization of the inertia of the layout (I), minimization of the empty volume (V) or minimization of the number of used bins/containers (B). If the layout is 3-dimensional, the mention "3D" is added to the objective column. }
  \label{tab:biblio}
\end{table}
\newpage
Virtual force-based algorithms are well known algorithms that are used for instance in robotics to solve highly constrained problems \cite{tarkesh,Ji,Brambilla,Khatib}, In this paper, a virtual force based algorithm is proposed to tackle balanced circular bin packing problems.
To the best of our knowledge, virtual force systems-based algorithms have never been applied to packing problems and seem to be promising techniques in terms of constraint satisfaction especially for high dimensional balanced bin packing problems. 
The rest of the paper is organized as follows: Section \ref{sec:formulation} introduces the mathematical formulation of the problem to solve. In Section \ref{sec:algo}, the proposed algorithm is detailed, and the conducted experiments as well as the corresponding results are reported, analyzed and compared with the methods in the literature on the same benchmark problems in Section \ref{sec:exp}.

\section{Formulation of the problem}
\label{sec:formulation}

The two-dimensional balanced circular bin packing problem considered in this paper is formulated as follows: pack a set of non-overlapping circles with given radii and masses, in the smallest circular container as possible, such that the center of gravity of the circles is located at the geometrical center of the container \cite{Xiao,Tang99}. Let $C_i, i\in\{1,N\}$ be a set of $N$ circles of radius $r_i^c$ and mass $m_i$. Each circle $C_i$ is located thanks to the coordinates of its center of inertia $(x_i,y_i)$. The vectors $\mathbf{x}$ and $\mathbf{y}$ encompass the 2D cartesian coordinates of all the circles.
The balanced circular bin packing problem can be mathematically formulated as follows:
\begin{equation}
\begin{aligned}
\min_{\textbf{x},\textbf{y}} \quad & r(\textbf{x},\textbf{y})=\max_{i\in \{1,N\}} \left(\sqrt{x_i^2+y_i^2}+r_i^c\right) \\
\textrm{where} \quad & \textbf{x}=\{x_1,...,x_N\} \in \mathbb{R}^{N}, \textbf{y}=\{y_1,...,y_N\} \in \mathbb{R}^{N}\\
\textrm{s.t.} \quad & \mathbf{h}_{overlap}(\mathbf{x,y}) = 0 \\
  &\mathbf{h}_{CG}(\mathbf{x,y}) = 0   \\
\end{aligned}
\label{eq:eq1}
\end{equation}

where $r$ corresponds to the objective function \textit{i.e.} minimize the container radius, $\mathbf{x}, \mathbf{y}$ are the vectors of continuous coordinates of the circles  and $\mathbf{h}_{overlap}(x,y)$, $\mathbf{h}_{CG}(x,y)$ are respectively the overlapping and balancing constraints with CG the center of gravity of the circles. 
The constraints are mathematically expressed as:
\begin{equation}
    \mathbf{h}_{overlap}(\mathbf{x,y})=\sum_{i=1}^{N-1} \sum_{j=i+1}^N \Delta S_{ij}(x_i,x_j,y_i,y_j)
\end{equation}
where $\Delta S_{ij}$ is the area of intersection of two circles $i$ and $j$.

\begin{equation}
    \mathbf{h}_{CG}(\mathbf{x,y})=\sqrt{\left(\frac{\sum_{i=1}^N m_ix_i}{\sum_{i=1}^Nm_i}\right)^2+\left(\frac{\sum_{i=1}^N m_iy_i}{\sum_{i=1}^Nm_i}\right)^2}
    \label{eq:cg}
\end{equation}

To sum up, the problem is to solve is a $2N$-dimensional continuous, single-objective and constrained optimization problem.

\section{Swarm Intelligence algorithm for balanced circular bin packing problems}
\label{sec:algo}
The proposed approach is a component swarm optimization algorithm based on a virtual-force system (CSO-VF), adapted from \cite{Gamot2} for solving the balanced circular bin packing problem.  Similar approaches are mainly used in the field of robotics in order, for instance, to control a swarm of robots moving in a given area with one or several objectives (\textit{e.g} reaching a target destination), as well as hard constraints (\textit{e.g.} avoid collisions and forbidden zones) \cite{Brambilla,Khatib}. This strategy is adapted in this paper in order to optimize the layout of various items in a container and is in line with quasi-physical models developed for instance in \cite{He,Liu5}. The main focus of the approach is to define dedicated operators for the evolution of a dynamical system of the components (\textit{i.e.} the circles) based on the fundamental principle of dynamics to satisfy the constraints and minimize the objective function in order to ensure an efficient resolution of the constraints as well as optimization capabilities. In this algorithm, each component is assumed to be a particle in a swarm. At each iteration of the algorithm, depending of the virtual forces that are applied to the particle, each of them evolves in the container in order to minimize the objective function and solve the constraints. It is important to note that this type of algorithm is different from classical particle swarm optimization. Indeed, in the CSO-VF algorithm, each particle of the swarm corresponds to a single circle and so to a part of the entire solution while in classical particle swarm optimization a particle corresponds to an entire solution.

\subsection{The virtual-force system}
\begin{figure}[h]
     \centering
     \begin{subfigure}[b]{0.45\textwidth}
         \centering
         \includegraphics[width=0.95\textwidth]{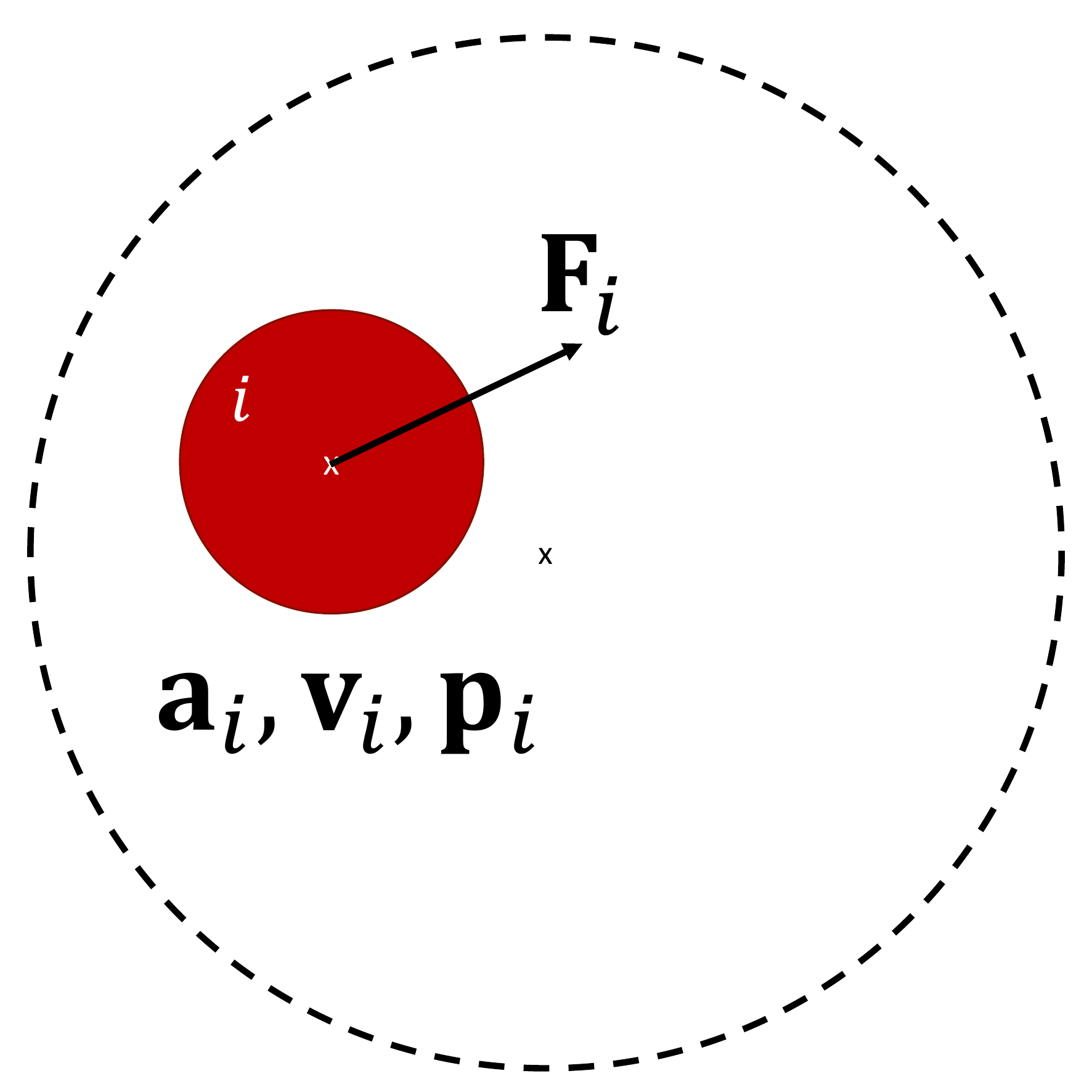}
         \caption{Definition of a circle.}
         \label{fig:circle}
     \end{subfigure}
     \hfill
     \begin{subfigure}[b]{0.45\textwidth}
         \centering
         \includegraphics[width=\textwidth]{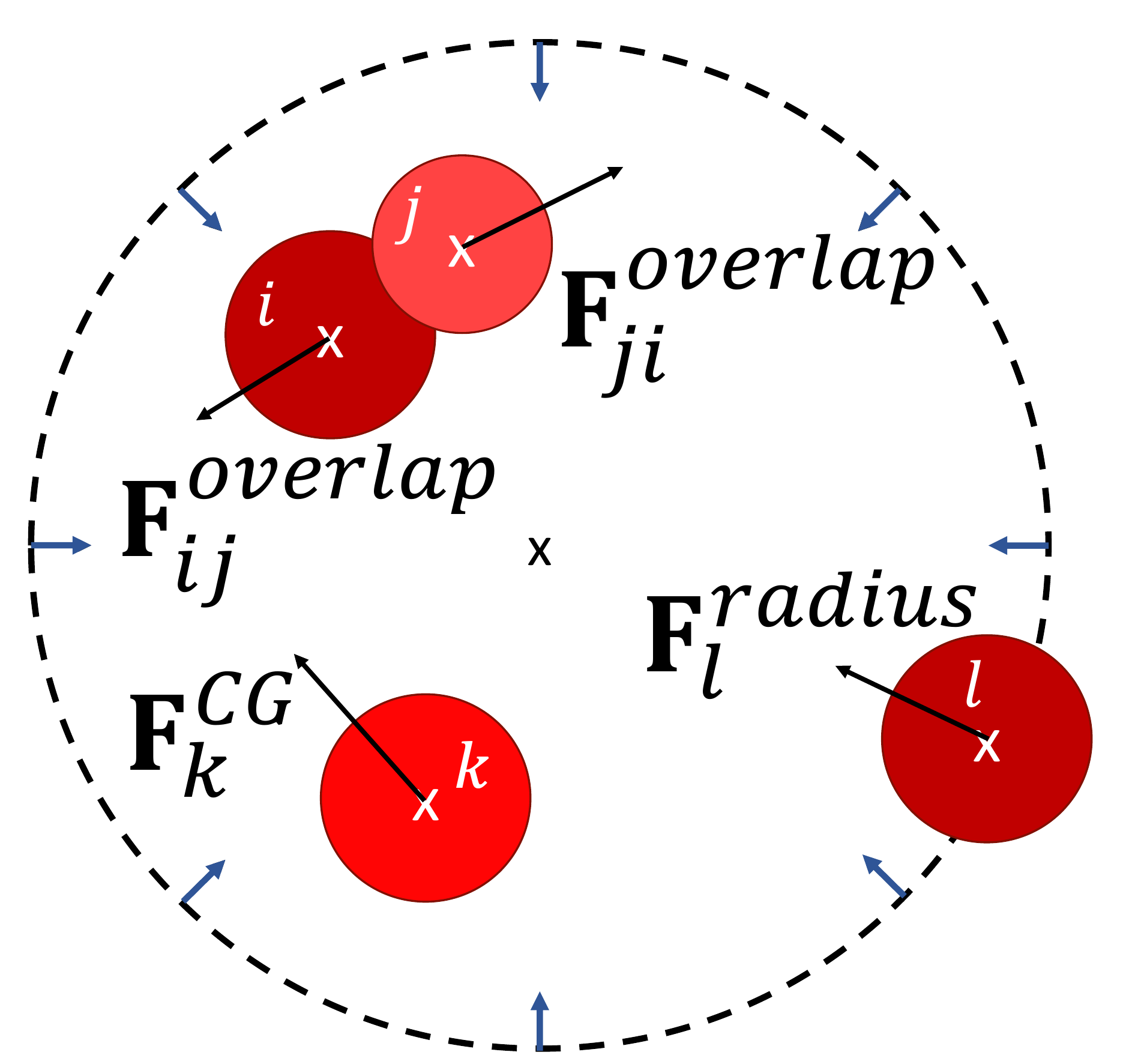}
         \caption{Definition of the virtual-force system for four circles $i,j,k$ and $l$.}
         \label{fig:forces}
     \end{subfigure}
     
        \caption{Definition of the circles' dynamic features and the associated virtual-force system.}
        \label{fig:def}
\end{figure}

In the CSO-VF algorithm, each circle $i$ is described by its dynamic features: its acceleration $\mathbf{a_i}$, its  speed $\mathbf{v}_i$ and its position $\mathbf{p}_i=\{x_i,y_i\}$. Forces ($\mathbf{F}_i^k, k\in\{1,...,N_F\}$, for circle $i$) with a resultant $\mathbf{F}_i$ are applied to the circle in order to move it at each iteration as illustrated on Figure \ref{fig:circle}. Each of the forces of the virtual-force system aims at solving the constraints as well as optimizing the objective function. Consequently, the forces $\mathbf{F}_i^k$ applied to the circle $i$ are of three types: $\mathbf{F}^{overlap}_{ij}$ to solve the overlapping constraints between the circles $i$ and $j$, $\mathbf{F}^{CG}_{i}$ to solve the balancing constraint and $\mathbf{F}^{radius}_{i}$ to minimize the objective function.

By the end of an iteration, the resulting force $\mathbf{F}_i$ is calculated for each circle thanks to Equation \ref{eq:resultant} and the fundamental principle of dynamics is applied in order to update the position of the swarm of circles at step $t+1$ (separated from step $t$ by $\Delta t$ which corresponds to a time unit) according to Equations \ref{eq:pfd1}, \ref{eq:pfd2} and \ref{eq:pfd3} (detailed for circle $i$).

\begin{equation}
\label{eq:resultant}
 \mathbf{F}_{i}=\left \{
  \begin{aligned}
    &\sum^{N_{F}}_{k=1} \mathbf{F}_i^k && \text{if}\ \left\lVert \sum^{N_{F}}_{k=1} \mathbf{F}_i^k \right\rVert  < F_{max} \\
    &\frac{\sum^{N_{F}}_{k=1} \mathbf{F}_i^k}{\left\lVert \sum^{N_{F}}_{k=1} \mathbf{F}_i^k \right\rVert} F_{max} && \text{otherwise.}
  \end{aligned} \right.
\end{equation} 

\vspace{5pt}
where $F_{max}$ is a hyperparameter corresponding to the maximum value of the norm of the resulting force vector.

\begin{equation}
\label{eq:pfd1}
    \mathbf{a}_{i,t+1} = \frac{\mathbf{F}_i}{m_i}
\end{equation}
\begin{equation}
\label{eq:pfd2}
    \mathbf{v}_{i,t+1} = \mathbf{v}_{i,t} + \mathbf{a}_{i,t+1} \Delta t
\end{equation}
\begin{equation}
\label{eq:pfd3}
    \mathbf{p}_{i,t+1} = \mathbf{p}_{i,t} + \mathbf{v}_{i,t+1} \Delta t
\end{equation}

\vspace{10pt}

The three forces of the virtual-force system are detailed and formulated as follows (for a circle $i$):
\begin{itemize}
    
    \item \textbf{The overlap constraint forces:} If two circles $i$ and $j$ are overlapping each other, repulsive forces $\mathbf{F}^{overlap}_{ij}$ and $\mathbf{F}^{overlap}_{ji}$ are applied to each of them as illustrated on Figure \ref{fig:forces}. The overlap forces are expressed as:
    \begin{equation}
 \mathbf{F}^{overlap}_{ij}=\left \{
  \begin{aligned}
    &-\frac{\mathbf{p}_j-\mathbf{p}_i}{\lVert\mathbf{p}_j-\mathbf{p}_i\rVert+\epsilon}v_{max}-\mathbf{v}_i && \text{if}\ \Delta S_{ij}(\mathbf{p}_i,\mathbf{p}_j) \ne 0 \\
    &\mathbf{0} && \text{otherwise.}
  \end{aligned} \right.
\end{equation} 

    where $v_{max}$ is a hyperparameter corresponding to the maximum value of the norm of the speed vector, $\epsilon$ ensures numerical stability and $\mathbf{0}=(0,0)$ the null vector.
    \item \textbf{The balance constraint forces:} In order to position the center of mass of the circles on the geometrical center of the container, gradient-based forces are applied along the opposite of the gradient of the position of the global center of mass according to the position of the center of inertia of each circle. This force is named $\mathbf{F}^{CG}_{i}$. It is illustrated on Figure \ref{fig:forces} with circle $k$.
    The balancing forces are expressed as:
    \begin{equation}
        \mathbf{F}^{CG}_i=-\alpha \nabla \mathbf{h}_{CG}(\mathbf{p}_i)
    \end{equation}
    where $\alpha$ is a "step-size" hyperparameter of the algorithm and $\nabla \mathbf{h}_{CG}(\mathbf{p}_i)$ corresponds to the gradient of the position of the center of gravity of the circles with respect to the position of the considered circle $i$. This type of force is inspired from gradient-based descent algorithms.
    
    \item \textbf{The objective function forces:} To minimize the radius of the container, the container is initialized to a given radius which decreases along with the iterations as described in Section \ref{sec:container}. The circles must be contained by the container and thus, an attractive force directed toward the geometrical center of the container is applied to each of the circles which does not belong to the container. This force is named $\mathbf{F}^{radius}_{i}$ and is illustrated on Figure \ref{fig:forces} with circle $l$. The radius forces are expressed as:
        \begin{equation}
 \mathbf{F}^{radius}_i =\left \{
  \begin{aligned}
    &\frac{\mathbf{p}_{c}-\mathbf{p}_i}{\lVert \mathbf{p}_{c}-\mathbf{p}_i \rVert+\epsilon}v_{max}-\mathbf{v}_i && \text{if}\ \Delta S_{i,container}(\mathbf{p}_c,\mathbf{p}_i) < S_i \\
    &\mathbf{0} && \text{otherwise.}
  \end{aligned} \right.
\end{equation} 
    
where $\Delta S_{i,container}$ corresponds to the area of intersection of circle $i$ with the container, $S_i$ to the area of circle $i$, $\mathbf{p}_{c}=(0,0)$ corresponds to the position's vector of the geometrical center of the container and $\epsilon$ ensures numerical stability.
\end{itemize}

\subsection{The evolution of the container's radius}
\label{sec:container}
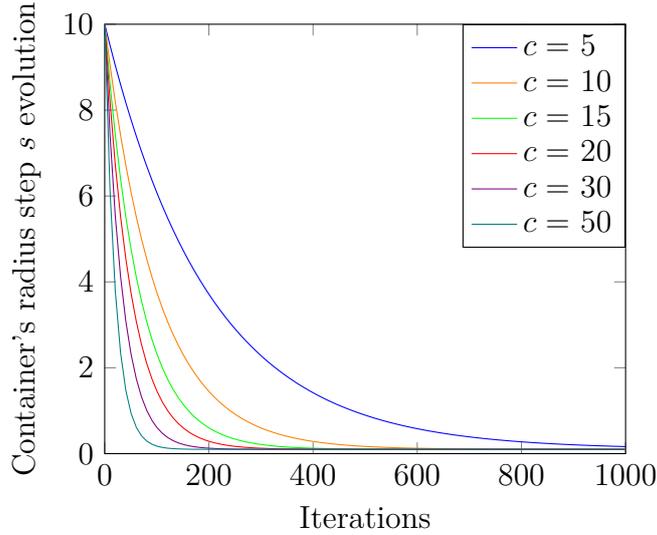
\begin{figure}[h!]
\centering
  \begin{tikzpicture}
\begin{axis}[ymin=0,xmin=0,ymax=10,xmax=1000,cycle list name = fcycle,xlabel={Iterations},xticklabel style={/pgf/number format/1000 sep=}, ylabel={Container's radius step $s$ evolution},legend cell align={left},
		legend style={at={(1,1)},anchor=north east},] 
]]
		
	\foreach \off in {5,10,15,20,30,50}{	% Tracé de la première fonction
		\addplot+[mark=none] expression[domain=0:1000,samples=100]{0.1+(10-0.1)*exp(x*ln(1/(1+\off/1000)))};
		\edef\VariableTempo{$c = $ \off}
		\expandafter\addlegendentry\expandafter{\VariableTempo}
	}
    \end{axis}
\end{tikzpicture}
  \caption{Influence of the hyperparameter $c$ on the evolution of the container's radius step size along with the iterations for: 1000 iterations, $s_{max}=10, s_{min}=0.1$.}
  \label{fig:c}
\end{figure}

In order to minimize the radius of the container and consequently to maximize the occupation rate of the circles in the container, the radius of the container is initialized as described in Section \ref{sec:init} and then decreases along with the iterations. Indeed, at each iteration (\textit{i.e.} timestep in the evolution of the dynamical system), if a feasible solution has been found (\textit{i.e.} all the constraints are satisfied), the current radius of the container $r_{t-1}$ is updated to a target container radius $r_{t}^{*}$ (where * stands for target), until a new feasible solution is found and the new actual container radius $r_t$ corresponding to the enclosing circle of the items centered on their center of gravity is calculated, and so on. The container is centered on the center of gravity of the circles to ensure that the center of gravity constraint (Equation \ref{eq:cg}) is automatically satisfied.
The proposed law of evolution of the target container radius $r_{t}^{*}$ is expressed as follows: 
\begin{equation}
\label{eq:rad1}
    r_{t}^{*}=r_{t-1}-s_{t} \\
\end{equation}
\begin{equation}
\label{eq:rad2}
    s_{t}=s_{min}+(s_{max}-s_{min})\exp \left( t\ln \left(\frac{1}{1+\frac{c}{N_{it}}}\right) \right)
\end{equation}
where $N_{it}$ is the maximum number of iterations, $s_{min}$ and $s_{max}$ are respectively the minimum and maximum radius steps, and $c$ is an hyperparameter characterizing the speed of evolution of the container radius along with the iterations. Figure \ref{fig:c} shows the influence of the hyperparameter $c$ on the law of evolution of the step $s_t$ along with the number of iterations.   

Moreover, if a stagnation of the convergence of the container's radius is observed during at least 10\% of the total number of iterations (\textit{i.e.} no feasible solution is found, at least one constraint is not satisfied), $s_t$ is set to $s_{min}$. Indeed, a stagnation of the convergence of the container's radius might be due to a too large step of the container's radius with respect to the convergence step. 

\subsection{Initialization of the algorithm}
\label{sec:init}
\subsubsection{Initialization of the container}
The radius of the container is initialized such that the occupation rate of the circles is equal to 15\%. The occupation rate $O$ of a layout is defined as:
\begin{equation}
    O=\frac{\sum_{i=1}^N S_i}{S_{container}}
\end{equation}
where $S_i$ is the surface of the circle $i$ and $S_{container}$ the surface of the container.

\subsubsection{Initialization of the positions of the circles}
\begin{figure}[h!]
    \centering
    \includegraphics[width=14cm]{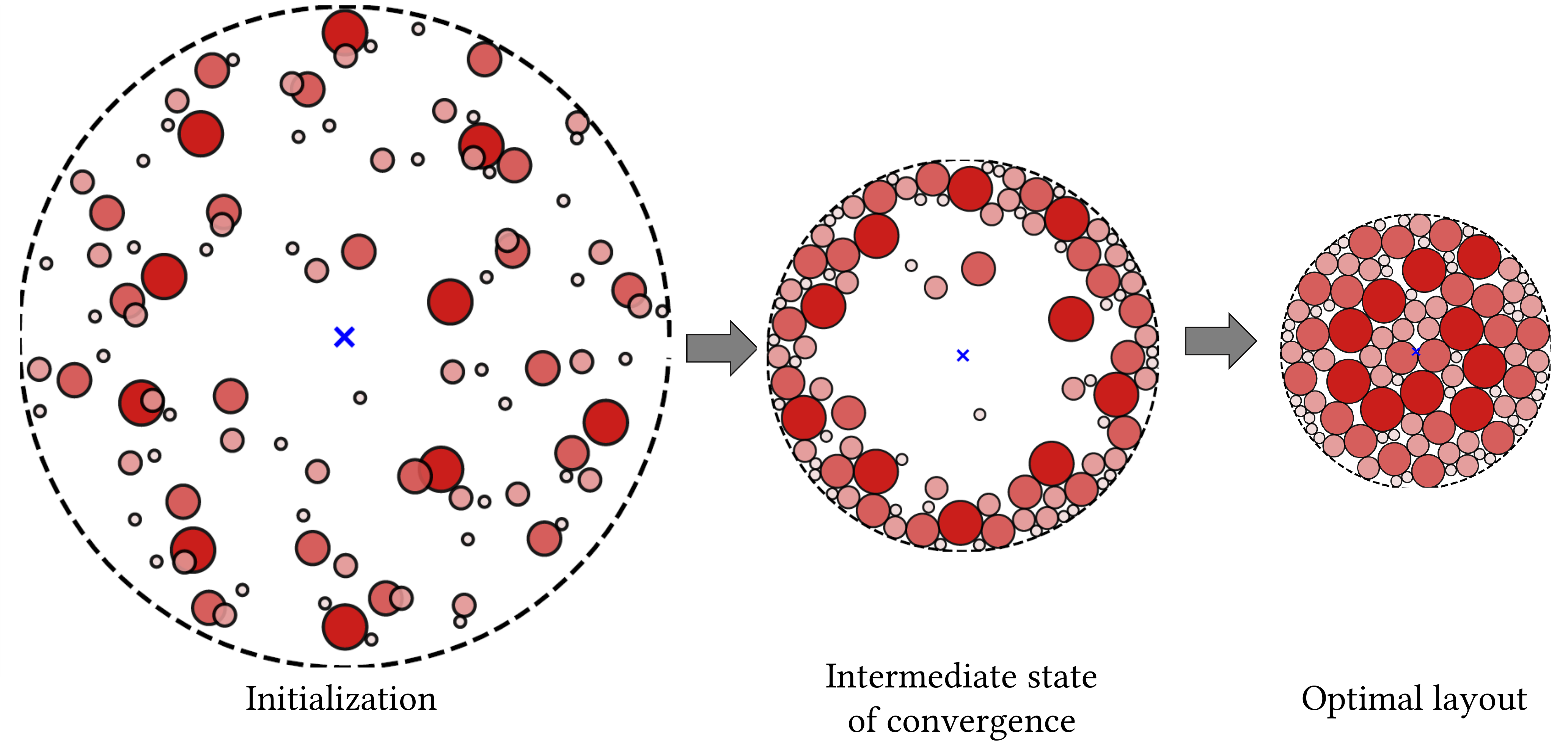}
    \caption{Evolution between an initialized container, an intermediate layout and the corresponding optimal layout found by the CSO-VF algorithm. The blue cross stands for the center of gravity of the circles. The black dotted circle stands for the container, centered on the center of gravity of the circles. The redder the circles, the heavier their corresponding masses.}
    \label{fig:init}
\end{figure}

The initialization of the circles is computed in two steps: The circles are grouped according to their masses in order to be positioned separately and each group is then positioned in the container thanks to a Latin Hypercube Sampling. This promotes an homogeneous distribution of the circles in the container according to their masses. Figure \ref{fig:init} shows the evolution between an initialized layout of the container, an intermediate layout and the corresponding optimal layout obtained with the CSO-VF algorithm for 100 unequal circles. The redder the circles, the heavier their corresponding masses. 

\subsection{The hyperparameters}
The 6 hyperparameters of the proposed CSO-VF algorithm are:
\begin{itemize}
    \item Dynamic features hyperparameters: maximum value of the norm of the force vector $F_{max}$ and speed vector $v_{max}$.
    \item Balancing force hyperparameter: $\alpha$.
    \item Radius law hyperparameters (in Equations \ref{eq:rad1}, \ref{eq:rad2}): maximum and minimum radius steps $s_{max}$ and $s_{min}$, speed evolution hyperparameter $c$.
\end{itemize}

\begin{figure}[h!]
    \centering
    \includegraphics[width=7.8cm]{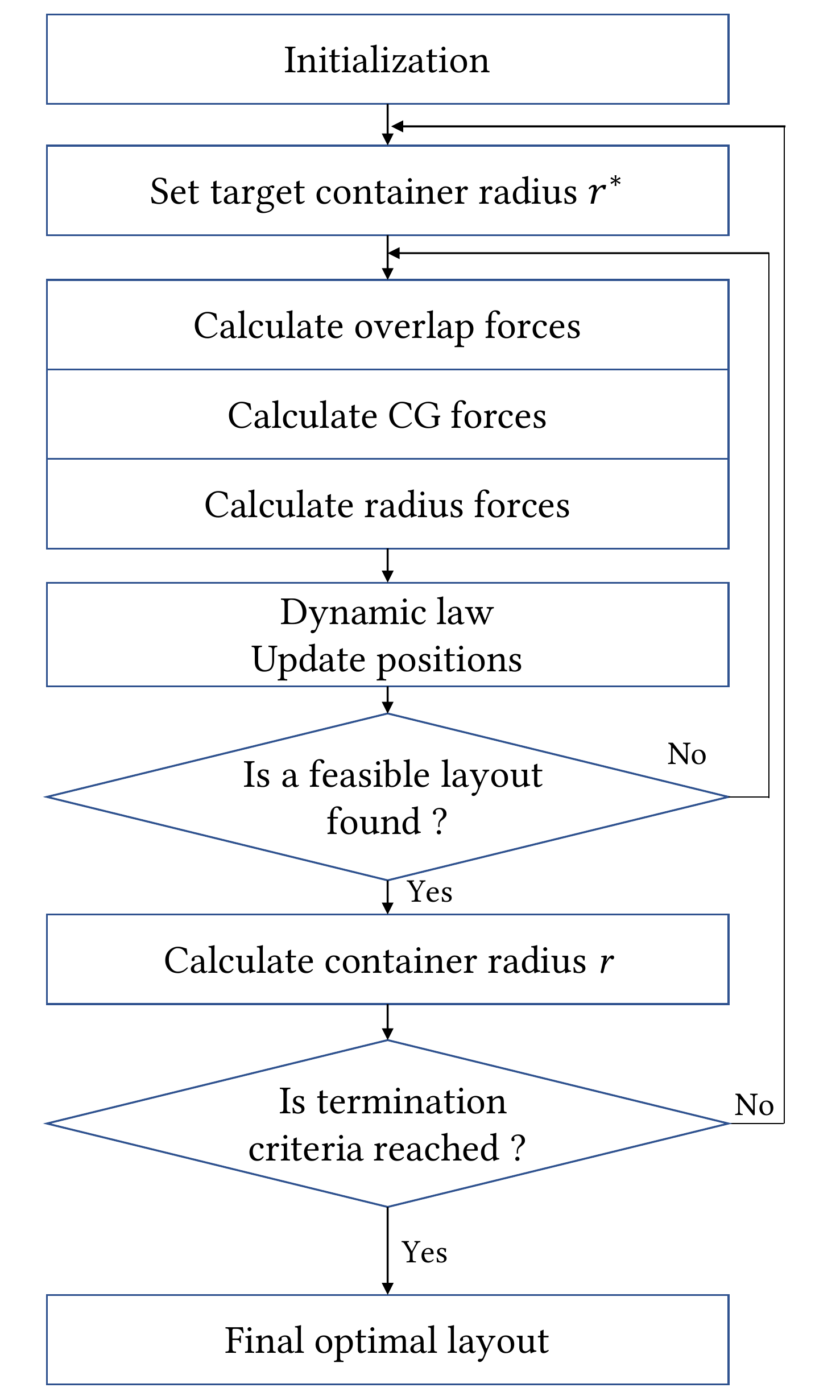}
    \caption{Steps of the CSO-VF algorithm.}
    \label{fig:pseudo}
\end{figure}

Figure \ref{fig:pseudo} describes the CSO-VF algorithm.\\

\section{Experimentations and results}
\label{sec:exp}
In this section, the CSO-VF algorithm is applied to two benchmarks of balanced circular bin packing problems of growing complexity, with respectively 10 and 3 problems taken from \cite{Xiao} and \cite{Romanova2}. The best obtained results are shown and analyzed, and a complementary study on the robustness of the CSO-VF algorithm is conducted on the most complex benchmark of problems.
\subsection{First benchmark of balanced circular bin packing problems}
\label{sec:inst1}
The first benchmark of balanced circular bin packing is initially described in \cite{Xiao}. This consists of 10 different sets of balanced circular bin packing problems from 10 to 55 circles of various sizes and masses, fully reported in Table \ref{tab:ap1} in \ref{ap:ap1}. Table \ref{tab:dim} sums up the maximal and minimal radii and masses of the circles for each problem of the benchmark. \\

\begin{table}[h!]
    \centering
    \begin{tabular}{|c|c|c|c|}
    \hline
        Pb. & Number of circles & Radii & Masses \\
        \hline
        \hline
         1 & 10 & 8-23 & 20-93 \\
         \hline
         2 & 15 & 8-24 & 12-98 \\
         \hline
         3 & 20 & 8-24 & 11-94 \\
         \hline
         4 & 25 & 6-24 & 11-96\\
         \hline
         5 & 30 & 6-24 & 12-96 \\
         \hline
         6 & 35 & 7-24 & 12-99 \\
         \hline
         7 & 40 & 6-23 & 12-99 \\
         \hline
         8 & 45 & 6-24 & 11-99 \\
         \hline
         9 & 50 & 5-24 & 11-99 \\
         \hline
         10 & 55 & 6-24 & 13-99 \\
         \hline
    \end{tabular}
    \caption{Minimum and maximum radii and masses of the circles for each problem of the benchmark (Pb.). They are detailed in \ref{ap:ap1}. }
    \label{tab:dim}
\end{table}

\begin{figure}[h!]
    \centering
    \begin{tikzpicture}
\begin{axis}[legend pos=north west,scaled y ticks=false,yticklabel style={/pgf/number format/1000 sep=},ymin=0,ymax=30000,xmin=0,xmax=11,xtick={1,2,3,4,5,6,7,8,9,10},
    ytick={0,5000,10000,15000,20000,25000}, xlabel={Benchmark problems},ylabel={Iterations}, ylabel style={yshift=10pt},  width=0.98\textwidth, height=0.5\textwidth,grid= major]
    \addplot[
        scatter,only marks,scatter src=explicit symbolic,
        scatter/classes={
            e={violet},
            d={BrickRed},
            c={ForestGreen},
            b={BurntOrange},
            a={RoyalBlue}
        }
    ]
    table[x=x,y=y,meta=label]{
        x    y    label
        1  3537 a
        1  3660 b
        1  4080 c
        1  4096 d
        1  4213 e
        2  2818 a
        2  3872 b
        2  7601 c
        2  13727 d
        2  19135 e
        3  4279 a
        3  5343 b
        3  10265 c 
        3  12077 d
        3  17001 e
        4  2061 a
        4  2859 b
        4  10668 c
        4  11848 d
        4  12426 e
        5  1394 a
        5  2052 b
        5  3507 c
        5  16309 d
        5  21223 e
        6  2096 a
        6  2576 b
        6  6373 c
        6  11027 d
        6  19930 e
        7  1637 a
        7  2070 b
        7  14542 c
        7  22996 d
        7  27649 e
        8  2608 a
        8  3743 b
        8  16036 c
        8  21165 d
        8  27769 e
        9  2745 a
        9  4234 b
        9  11464 c
        9  24181 d
        9  29125 e
        10  2688 a
        10  4193 b
        10  6404 c
        10  11080 d
        10  16715 e
    };
    %\draw[xstep=1,ystep=1000,very thin, color=gray] (0,0) grid (10,20000);
    \legend{0.1\%,0.5\%,1\%,5\%,10\%}
\end{axis}
\end{tikzpicture}
    \caption{Number of iterations to reach 10\%, 5\%, 1\%, 0.5\% and 0.1\% of the final obtained values for the best runs of the 10 benchmark problems.}
    \label{fig:steps1}
\end{figure}
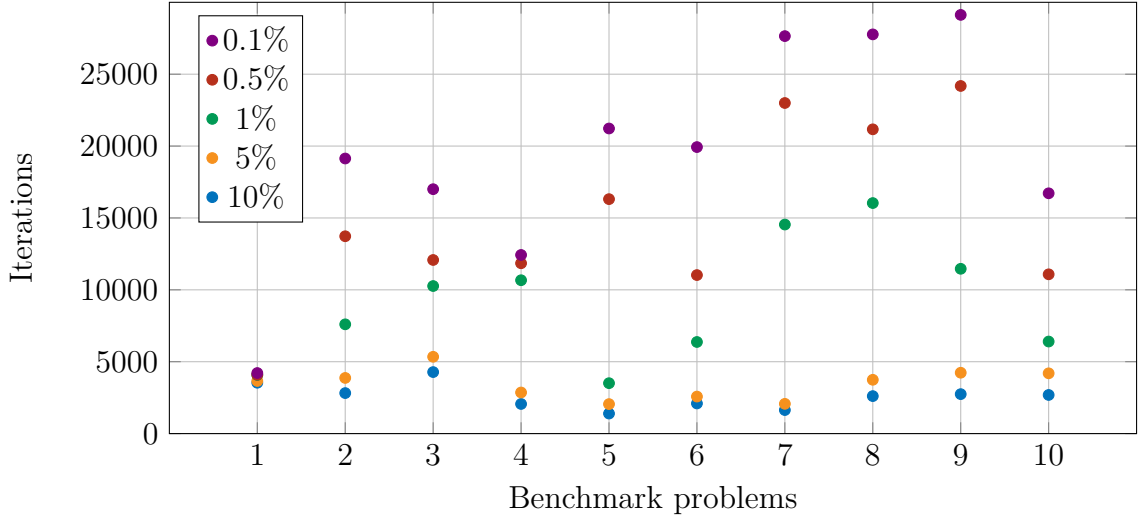

\begin{figure}[h!]
    \centering
    \includegraphics[width=15cm]{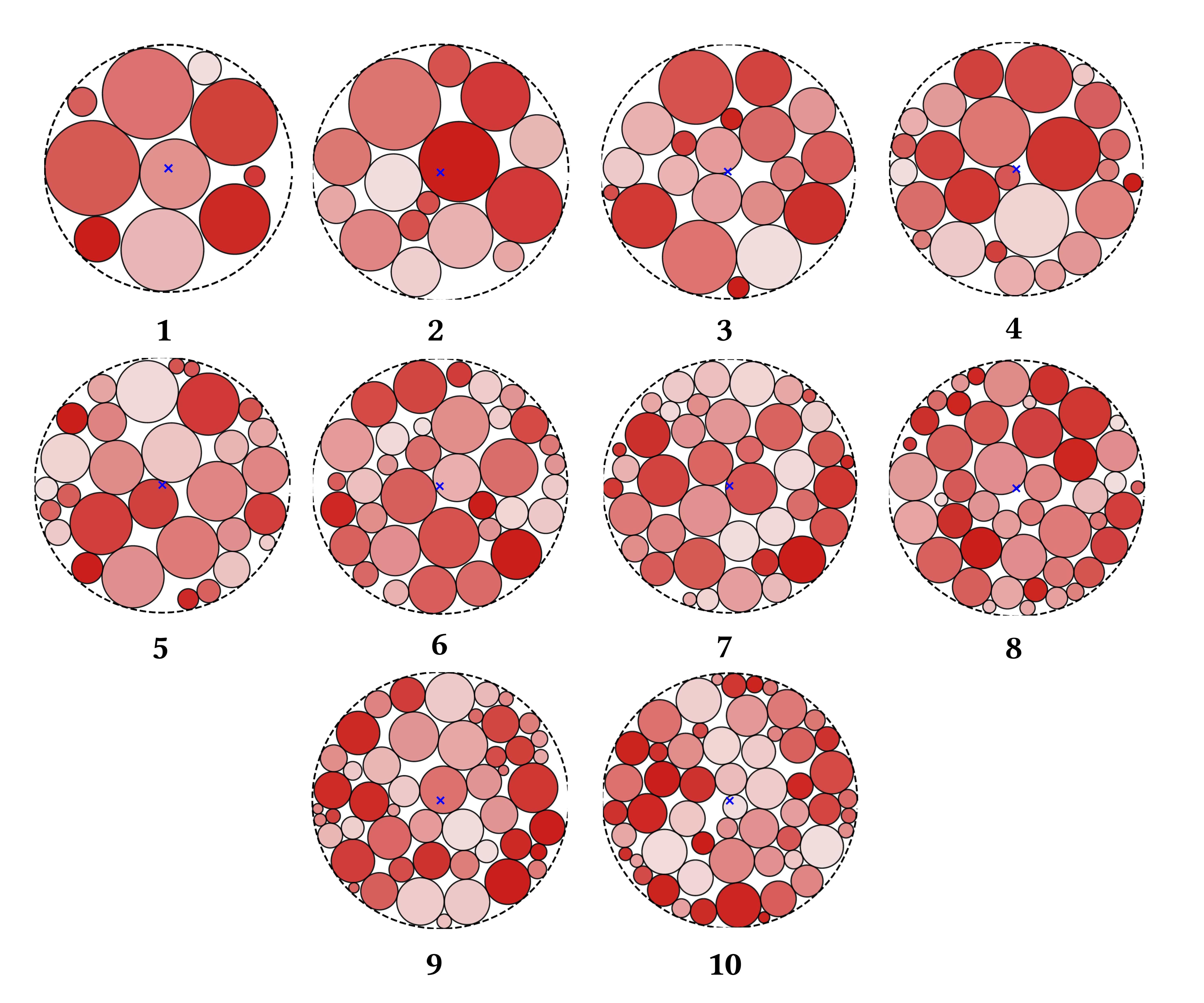}
    \caption{Obtained final best layout for the first benchmark of problems with the CSO-VF algorithm.}
    \label{fig:res1}
\end{figure}

The CSO-VF algorithm is applied to the 10 problems of this first benchmark with 100 repetitions each from different initializations using the process described in Section \ref{sec:init}. The hyperparameters are set thanks to a parametric study. \\
The results are outlined as follows:
\begin{itemize}
    \item The best obtained layouts' radii are reported on Table \ref{tab:res1} and compared with those published in \cite{Xiao} obtained with the PSO algorithm described in \cite{zhou}, those obtained in \cite{Xiao} with a compaction algorithm based on gradient search and SA (Comp.+SA), those obtained in \cite{Xiao} with a compaction algorithm based on gradient search and PSO (Comp.+PSO) and those from \cite{Liu4} obtained with a combination of the energy landscape paving method and local search procedure (IELP-LS).
    \item Figure \ref{fig:steps1} characterizes the speed of convergence: it shows the number of iterations necessary to reach 10\%, 5\%, 1\%, 0.5\% and 0.1\% of the final value for the best run of all the 10 problems.
    
    \item Figure \ref{fig:res1} shows the best obtained layout for each of the 10 problems. The redder the circles, the heavier they are.
    \item In \ref{ap:time1}, Table \ref{tab:time1} reports the CPU time necessary to run 20000 iterations with the allocated numerical resources for each benchmark problem.
\end{itemize}

\begin{table}[h!]
\centering
  \begin{tabularx}{1.0\textwidth}{ 
  | >{\centering\arraybackslash}m{0.95cm}
  | >{\centering\arraybackslash}m{1.75cm} 
  | >{\centering\arraybackslash}m{1.75cm} 
  | >{\centering\arraybackslash}m{1.75cm} 
  | >{\centering\arraybackslash}m{1.75cm} 
  || >{\centering\arraybackslash}m{1.75cm} 
  | >{\centering\arraybackslash}m{1.75cm} |}
    \hline
    Pb. & PSO \cite{zhou} & Comp. + SA \cite{Xiao} & Comp. + PSO \cite{Xiao} &  IELP-LS \cite{Liu4} & Proposed CSO-VF algorithm & Relative Improvement (\%) wrt best method  \tabularnewline
    \hline
    1 & 61.32 & 60.96 & 59.93 & 59.92 &  \textbf{59.85} & 0.12\%    \\
    \hline
    2 & 76.58 &  68.77 & 67.65 & 67.39 & \textbf{67.07} & 0.47\%   \\
    \hline
    3 & 89.15 & 83.09 & 83.06 & 82.99 & \textbf{82.58} & 0.49\%   \\
    \hline
    4 & 106.31 & 83.97 & 84.24 & 82.98 &  \textbf{82.84} & 0.17\%   \\
    \hline
    5 & 136.88 & 99.58 & 99.89 & 98.97 & \textbf{98.77} & 0.20\%  \\
    \hline
    6 & 148.39 & 102.86 & 102.71 & 102.32 &   \textbf{101.52} & 0.78\%   \\
    \hline
    7 & 165.79 & 115.15 & 115.58 & 115.00 &  \textbf{113.53} & 1.28\% \\
    \hline
    8 & 172.69 & 120.63 & 119.67 & 119.07 &  \textbf{117.99} & 0.90\%  \\
    \hline
    9 & 189.89 & 125.82 & 126.19 & 124.98 &  \textbf{124.30} & 0.54\%    \\
    \hline
    10 & 200.82 & 138.22 & 138.89 & 136.13 &  \textbf{135.99} & 0.10\%   \\
    \hline
  \end{tabularx}
  \caption{Results for the first benchmark of 10 problems (Pb.): best container's radius obtained in \cite{zhou,Xiao,Liu4} and with the CSO-VF algorithm. Best results are indicated in bold.}
  \label{tab:res1}
\end{table}

%\begin{figure}[h!]
%    \centering
%    \input{histo1.tex}
%    \caption{Number of runs (amongst 100) to reach 10\%, 5\%, 1\% and 0.5\% of the final obtained values (Table \ref{tab:res1}) for the best runs of the 10 problem instances.}
%    \label{fig:histo1}
%\end{figure}

The proposed strategy allows to solve all the benchmark problems. It enables to improve the final best radius of all the problems, from 0.1\% to 1.28\% in comparison to the results from the literature obtained with various techniques \cite{Xiao,Liu4,zhou}. This relative improvement tends to increase with the problem number (\textit{i.e.} with the number of circles) except for the last one. This might be explained by the fact that the CSO-VF algorithm is provided with dedicated forces to solve each of the constraints and thus, remains consistent in its ability to resolve constraints from 10 to 55 circles.\\
Figure \ref{fig:steps1} shows that for all 10 problems less than 5000 iterations are necessary to reach 5\% of the final best value. This represents from 43 seconds (problem 1, 10 circles) to 3.5 minutes (problem 10, 55 circles) thanks to Table \ref{tab:time1}. Figure \ref{fig:steps1} also highlights that the speed of convergence until 5\% of the final value seems not linked to the number of circles of the benchmark problems. However, problems 7, 8 and 9 (\textit{i.e.} 40, 45 and 50 circles) seem to require more iterations to converge to 0.1\% of their final value. \\

\subsection{Second benchmark of balanced circular bin packing problems}
The second benchmark of problems is initially proposed in \cite{Romanova2}. This consists of 3 problems with 100, 150 and 300 circles, in which the radius of each circles is equal to its mass. Table \ref{tab:inst2} lists the number of circles and dimensions for each problem of the benchmark.

\begin{table}[h!]
\centering
  \begin{tabular}{| >{\centering\arraybackslash}m{2cm}
  | >{\centering\arraybackslash}m{1.5cm}
  | >{\centering\arraybackslash}m{1.5cm}
  | >{\centering\arraybackslash}m{1.5cm}
  | >{\centering\arraybackslash}m{1.5cm}
  | >{\centering\arraybackslash}m{1.5cm}
  | >{\centering\arraybackslash}m{1.5cm}|}
    \hline
    \multirow{2}{*}{Pb.} &
      \multicolumn{2}{c|}{1} &
      \multicolumn{2}{c|}{2} &
      \multicolumn{2}{c|}{3} \\
      \vspace{10pt}
      
    & Number of circles & Radii, masses & Number of circles & Radii, masses &Number of circles & Radii, masses \\
    \hline
    \multirow{5}{*}{Circles}  & 40 & 10 & 50 & 10 & 100 & 10  \\& 30 & 20 & 40 & 20 & 80 & 20  \\& 20 & 30 & 30 & 30 & 60 & 30 \\ & 10 & 40 & 20 & 40 & 40 & 40 \\ &   &   & 10 & 50 & 20 & 50   \\
    \hline
    Total number of circles & \multicolumn{2}{c|}{100} & \multicolumn{2}{c|}{150} & \multicolumn{2}{c|}{300} \\
    \hline
    
  \end{tabular}
  \caption{Configuration of the 3 benchmark problems (Pb.): Number of circles, dimensions (radii equal to masses) and total number of circles for each problem.}
  \label{tab:inst2}
\end{table}

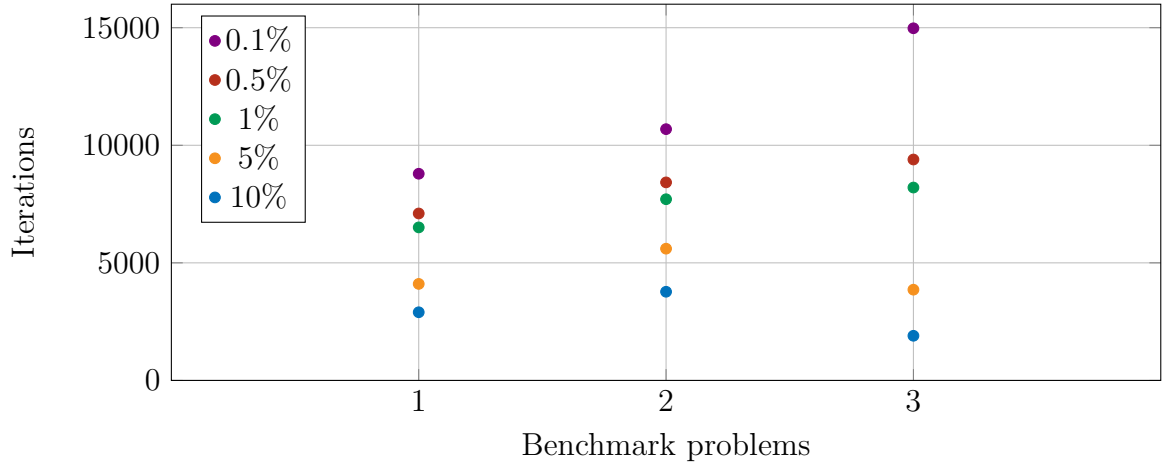
\begin{figure}[h!]
    \centering
    \begin{tikzpicture}
\begin{axis}[legend pos=north west,scaled y ticks=false,yticklabel style={/pgf/number format/1000 sep=},ymin=0,ymax=16000,xmin=0,xmax=4,xtick={1,2,3},
    ytick={0,5000,10000,15000}, xlabel={Benchmark problems},ylabel={Iterations}, ylabel style={yshift=10pt},  width=1\textwidth, height=0.45\textwidth,grid= major]
    \addplot[
        scatter,only marks,scatter src=explicit symbolic,
        scatter/classes={
            e={violet},
            d={BrickRed},
            c={ForestGreen},
            b={BurntOrange},
            a={RoyalBlue}
        }
    ]
    table[x=x,y=y,meta=label]{
        x    y    label
        1  2899 a
        1  4105 b
        1  6508 c
        1  7100 d
        1  8790 e
        2  3771 a
        2  5603 b
        2  7706 c
        2  8422 d
        2  10686 e
        3  1899 a
        3  3859 b
        3  8203 c 
        3  9395 d
        3  14976 e
        
    };
    %\draw[xstep=1,ystep=1000,very thin, color=gray] (0,0) grid (10,20000);
    \legend{0.1\%,0.5\%,1\%,5\%,10\%}
\end{axis}
\end{tikzpicture}
    \caption{Number of iterations to reach 10\%, 5\%, 1\%, 0.5\% and 0.1\% of the obtained value for the best run of the 3 problems of the benchmark.}
    \label{fig:steps2}
\end{figure}

\begin{figure}[h!]
    \centering
    \begin{tikzpicture}
\begin{axis}[
    %small,
    ybar,%=8pt, % configures ‘bar shift’
    enlargelimits=0.30,
    ylabel={Number of runs},
    xlabel={Benchmark problems},
    symbolic x coords={1, 2, 3},
    xtick={1,2,3},
    width=0.7\textwidth, height=0.45\textwidth,
    legend style={at={(0.5,-0.25)},
    anchor=north,legend columns=-1},
    nodes near coords,
    every node near coord/.append style={font=\tiny},
   nodes near coords align={vertical},
    ]
\addplot[style = {fill=RoyalBlue}] coordinates {(1, 100) (2,100) (3, 100) };
\addplot[style = {fill=BurntOrange}] coordinates {(1, 100) (2, 100) (3, 100) };
\addplot[style = {fill=ForestGreen}] coordinates {(1, 45) (2, 64) (3,100) };
\addplot[style = {fill=BrickRed}] coordinates {(1, 12) (2, 19) (3, 42) };
\legend{10\%, 5\%, 1\%,0.5\%},

\end{axis}
\end{tikzpicture}
    \caption{Number of runs (amongst 100) to reach 10\%, 5\%, 1\% and 0.5\% of the obtained value for the best run of the 3 problem of the benchmark.}
    \label{fig:histo2}
\end{figure}
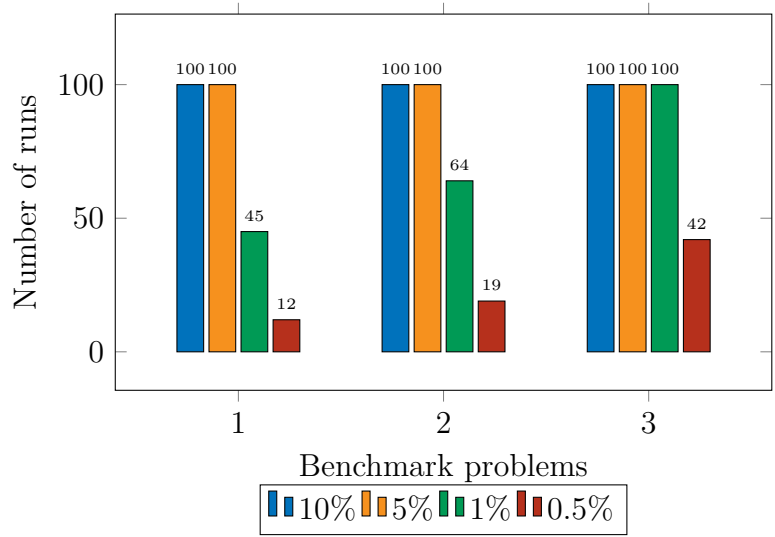

\begin{figure}[h!]
    \centering
    \includegraphics[width=15cm]{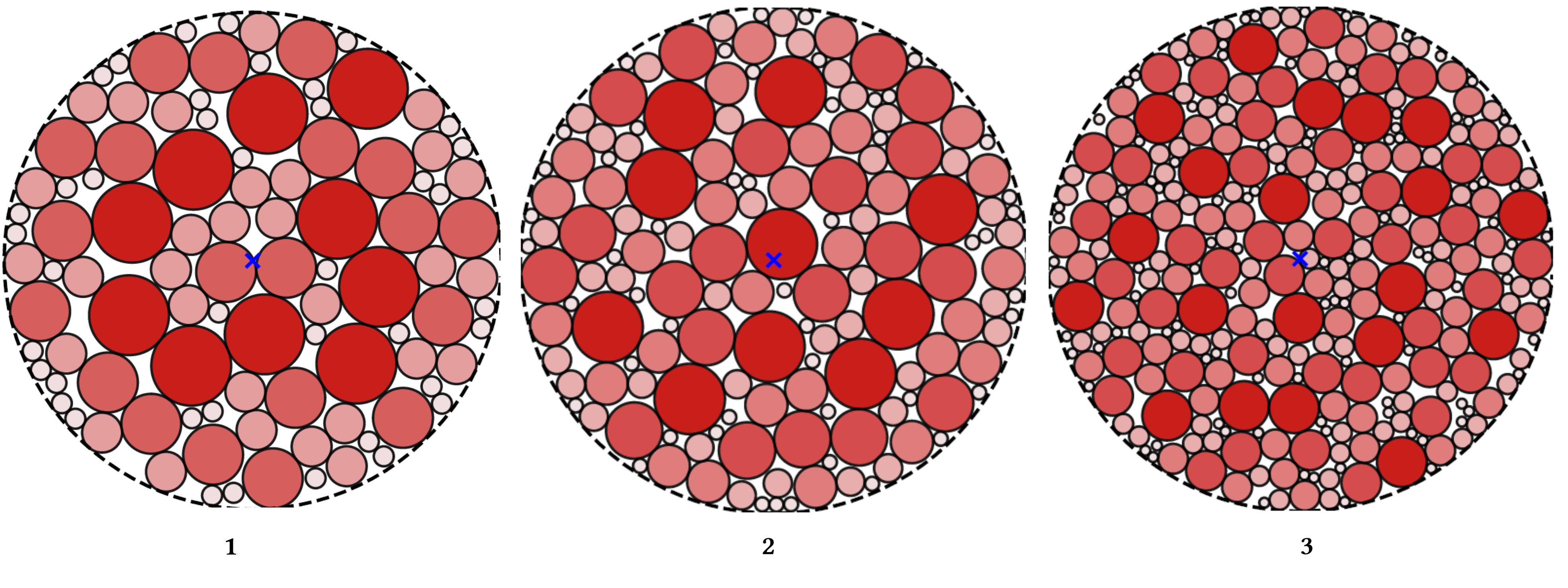}
    \caption{Obtained final best layout for the second benchmark of problems with the CSO-VF algorithm.}
    \label{fig:res2}
\end{figure}

The hyperparameters are set using a parametric analysis and the CSO-VF algorithm is applied to the 3 problems of this second benchmark with 100 repetitions each from different initializations using the process described in Section \ref{sec:init}. \\
The results are outlined as follows:
\begin{itemize}
    \item The best obtained layouts are reported on Table \ref{tab:res2} and compared with those obtained in \cite{Romanova2} with a sequential algorithm based on a variant of the Shor's r-algorithm \cite{shor,stetsyuk}.
    \item Figure \ref{fig:steps2} characterizes the speed of convergence: it shows the number of iterations necessary to reach 10\%, 5\%, 1\%, 0.5\% and 0.1\% of the final value for the best run of all the benchmark problems.
    \item Figure \ref{fig:histo2} characterizes the robustness of the algorithm on each problem: it shows the numbers of repetitions that reach 10\%, 5\%, 1\% and 0.5\% of the best layout for all the benchmark problems.
    \item Figure \ref{fig:res2} shows the best obtained layout for each problem of this benchmark. The redder the circles,  the heavier their corresponding masses.
    \item In \ref{ap:time2}, Table \ref{tab:time2} reports the CPU time necessary to run 15000 iterations with the allocated numerical resources for all the benchmark problems.
\end{itemize}

The results are reported in Table \ref{tab:res2} and compared with the results obtained in \cite{Romanova2}. Figure \ref{fig:res2} shows the best obtained layout for each problem of the benchmark.

\begin{table}[h!]
\centering
  \begin{tabularx}{1.0\textwidth}{ 
  | >{\centering\arraybackslash}m{2.35cm}
  | >{\centering\arraybackslash}m{3.5cm}
 | >{\centering\arraybackslash}m{3.5cm} 
  | >{\centering\arraybackslash}m{3.5cm} |}
    \hline
   Pb. & Results from \cite{Romanova2} & Proposed CSO-VF algorithm & Relative Improvement (\%)\\
   \hline
   1 & 257.19 & \textbf{247.93} & 3.60\%\\
   \hline
   2 & 368.40 & \textbf{357.97} & 2.83\% \\
   \hline
   3 & 520.56 & \textbf{504.11} & 3.16\% \\
   \hline
  \end{tabularx}
  \caption{Results for the second benchmark of problems (Pb.): best container's radius obtained in \cite{Romanova2} and with the CSO-VF algorithm.}
  \label{tab:res2}
\end{table}

Table \ref{fig:res2} and Figure \ref{fig:res2} show that the algorithm succeeds in finding feasible layouts for balanced circular bin packing problems up to 300 circles. It provides better container's radii from 2.83\% to 3.60\% compared to the best known layouts proposed in \cite{Romanova2}. Figure \ref{fig:steps2} shows that for all 3 problems of the benchmark, around 5000 iterations are necessary to reach 5\% of the final value. Moreover, Figure \ref{fig:steps2} shows that the number of iterations needed to reach less than 1\% of the final best value increases along with the total number of circles in each benchmark problem. However, for all three problems, 5\% of their best final value is reached in about 5000 iterations.
Figure \ref{fig:histo2} characterizes the robustness of the CSO-VF algorithm. It shows that for the 3 bin packing problems of the benchmark, all the runs manage to converge to at least 5\% of the final best layouts. Furthermore, the robustness of the algorithm increases along with the total number of circles of each problem. Indeed, for the 300-circles benchmark problem (\textit{i.e.} the third one), the CSO-VF algorithm achieves 100\% of repetitions reaching 1\% of the best radius obtained in Table \ref{tab:res2}, and more than 40\% of repetitions reaching 0.5\% of the best radius obtained in Table \ref{tab:res2}. 

\subsection{Summary of the results for the two benchmarks}
The CSO-VF algorithm provides better optimal layouts for all 13 problems where 10 to 300 circles had to be positioned in comparison with several techniques from the literature. It has been shown that the CSO-VF algorithm solves all the benchmarks problems with similar convergence speed. As the number of circles increases from 100 to 300 circles, the algorithm appears to be more robust and able to achieve up to 0.5\% of the best final value. It might be due to the fact that problems with fewer circles are more sensitive to the initialization than problems with several hundreds of circles and especially if they are uniform in terms of dimensions and masses. 
The virtual-force system of the CSO-VF algorithm uses dedicated operators to solve each of the constraints as well as minimize the objective function. Therefore, the CSO-VF algorithm provides constant and efficient optimization capabilities regardless of the considered number of circles.

\section{Conclusions and future works}

This paper describes a quasi-physical approach in order to solve balanced circular bin packing problems which aim is to pack weighted circles into as smallest balanced container as possible. The proposed algorithm consists in a swarm intelligence based on a virtual-force system which provides dedicated operators to solve each of the constraints as well as optimize the objective function. It has been successfully applied on various benchmark problems from 10 to 300 circles. The reported results show that this strategy allows to improve the results obtained for the considered benchmark problems compared to other various counterparts from the literature. 
In the future, the following perspectives will be investigated. Firstly, the evolution of the container's radius step law will be improved to increase the robustness of the method. Secondly, the algorithm will be extended in order to tackle 3-dimensional bin packing problems.

\newpage
\appendix
\section{First benchmark problems configurations}
\label{ap:ap1}
Table \ref{tab:ap1} details the dimensions of the circles of the first benchmark of problems addressed in Section \ref{sec:inst1}.

\begin{table}[h!]
\centering
\small
  \begin{tabularx}{1.0\textwidth}{ 
  | >{\centering\arraybackslash}m{0.6cm} 
  | >{\centering\arraybackslash}m{0.6cm} 
  | >{\centering\arraybackslash}m{5.65cm} 
  | >{\centering\arraybackslash}m{6cm}  |}
    \hline
    Inst & Size & Radii  & Masses\\
    \hline
    \hline
    I1 & 10  & \{20, 22, 17, 17, 7, 21, 11, 5, 23, 8\}& \{35, 61, 49, 89, 68, 80, 93, 82, 70, 20\} \\
    \hline
    I2 & 15  & \{8, 14, 8, 15, 11, 17, 21, 16, 6, 18, 24, 13, 20, 10, 15\}& \{75, 29, 36, 58, 75, 32, 98, 52, 76, 85, 59, 18, 85, 36, 12\} \\
    \hline
    I3 & 20  & \{20, 24, 8, 11, 13, 7, 7, 15, 24, 18, 15, 17, 17, 14, 16, 18, 5, 21, 21, 13\}& \{86, 72, 81, 54, 29, 94, 92, 41, 57, 77, 40, 67, 31, 47, 39, 61, 73, 83, 11, 20\} \\
    \hline
    I4 & 25  & \{24, 16, 19, 7, 14, 24, 15, 6, 16, 16, 23, 10, 9, 10, 18, 22, 7, 9, 7, 13, 14, 8, 18, 6, 8\}& \{16, 80, 52, 21, 42, 86, 67, 96, 61, 79, 57, 62, 32, 38, 20, 75, 80, 11, 53, 32, 41, 68, 85, 53, 71\} \\
    \hline
    I5 & 30  & \{14, 15, 11, 19, 9, 6, 23, 9, 23, 13, 24, 12, 24, 24, 10, 8, 9, 8, 6, 11, 6, 16, 24, 12, 9, 19, 13, 24, 21, 18\}& \{24, 52, 37, 17, 12, 19, 51, 67, 23, 46, 14, 96, 55, 84, 21, 92, 69, 65, 72, 36, 73, 83, 83, 97, 73, 81, 30, 46, 49, 51\} \\
    \hline
    I6 & 35  & \{10, 20, 13, 19, 19, 10, 14, 14, 24, 11, 20, 15, 7, 18, 22, 10, 13, 12, 21, 14, 9, 10, 9, 7, 8, 18, 8, 8, 23, 14, 13, 21, 23, 16, 10\}& \{44, 46, 14, 32, 70, 31, 95, 24, 75, 99, 99, 79, 10, 79, 69, 64, 12, 47, 41, 62, 17, 85, 43, 70, 43, 63, 44, 57, 62,
20, 17, 80, 47, 68, 19\} \\
\hline
    I7 & 40  & \{6, 12, 20, 6, 14, 19, 9, 20, 10, 13, 12, 14, 23, 17, 16, 19, 15, 10, 12, 18, 21, 6, 20, 17, 13, 20, 17, 6, 21, 15, 12, 9, 14,
20, 23, 16, 23, 9, 23, 18\}& \{74, 48, 16, 35, 19, 58, 87, 90, 17, 29, 32, 63, 46, 76, 26, 88, 71, 49, 89, 14, 68, 94, 41, 53, 36, 67, 14, 88, 99, 46, 66,
14, 21, 44, 73, 72, 72, 37, 82, 12\} \\
\hline
    I8 & 45  & \{13, 8, 11, 21, 9, 20, 24, 20, 17, 21, 7, 13, 24, 7, 6, 8, 18, 15, 12, 18, 17, 21, 8, 23, 22, 15, 10, 17, 24, 8, 14, 6, 16, 14, 6, 10,
19, 21, 20, 6, 16, 14, 6, 19, 11\}& \{91, 95, 96, 47, 63, 37, 56, 96, 84, 70, 36, 41, 48, 12, 86, 43, 70, 71, 56, 89, 52, 49, 53, 82, 42, 35, 11, 82, 88, 58, 74,
16, 91, 57, 26, 39, 48, 68, 72, 69, 27, 44, 25, 99, 96\} \\
\hline
    I9 & 50  & \{9, 17, 5, 15, 24, 23, 12, 9, 5, 13, 7, 18, 19, 21, 7, 18, 18, 24, 12, 23, 22, 13, 5, 6, 17, 21, 7, 18, 14, 17, 10, 15, 18, 8, 8, 16,
7, 18, 24, 6, 20, 10, 21, 11, 22, 24, 12, 7, 14, 11\}& \{19, 85, 60, 19, 88, 18, 28, 55, 66, 47, 49, 69, 93, 94, 35, 43, 93, 34, 27, 61, 20, 52, 51, 41, 98, 85, 82, 89,
54, 43, 54, 94, 80, 99, 41, 41, 63, 28, 19, 53, 11, 78, 65, 10, 98, 43, 78, 24, 84, 16\} \\
\hline
    I10 & 55  & \{17, 23, 17, 13, 18, 21, 23, 22, 7, 9, 8, 13, 20, 11, 10, 19, 10, 14, 12, 22, 19, 10, 17, 11, 21, 8, 15, 16, 19, 21, 17, 19, 8, 6,
13, 13, 14, 19, 18, 23, 20, 24, 24, 13, 13, 19, 7, 6, 10, 8, 8, 10, 24, 19, 24\}& \{97, 62, 28, 36, 97, 58, 13, 21, 40, 97, 79, 90, 62, 47, 64, 23, 23, 95, 99, 44, 71, 79, 52, 59, 47, 60, 41, 47, 90, 95, 81,98, 70, 47, 90, 13, 93, 50, 21, 80, 17, 52, 96, 73, 88, 16, 91, 97, 40, 52, 50, 90, 19, 69, 14\} \\

    \hline
  \end{tabularx}
  \caption{Geometrical configuration of each problem: total number of circles (size), their radii and masses. }
  \label{tab:ap1}
\end{table}

\newpage
\section{CPU time of resolution}
The routines are implemented in Python language, and executed on a PC with an Intel Core i7, 16 GB of RAM and Windows operating system. The CPU time of resolution for each benchmark problem is reported and analyzed in the following sections.
\subsection{First benchmark of problems}
\label{ap:time1}
\begin{table}[h!]
    \centering
    \begin{tabular}{|c|c|c|c|c|c|c|c|c|c|c|}
    \hline
         Pb. & 1 & 2 & 3 & 4 & 5 & 6 & 7 & 8 & 9 & 10 \\
         \hline
         Time (min) & 1.75 & 3.62 & 4.05 & 5.37 & 6.12 & 8.43 & 10.6 & 12.35 & 13.14 & 13.89\\ 
         \hline
    \end{tabular}
    \caption{Mean of CPU time for 20000 iterations for the ten benchmark problems (Pb.). Implementation: Python language, Execution: PC with an Intel Core i7, 16 GB of RAM, Operating system: Windows. }
    \label{tab:time1}
\end{table}
\subsection{Second benchmark problems}
\label{ap:time2}
\begin{table}[h!]
    \centering
    \begin{tabular}{|c|c|c|c|}
    \hline
         Benchmark Problems & 1 & 2 & 3  \\
         \hline
         CPU time (min) & 20.9 & 42.1 & 98.7 \\ 
         \hline
    \end{tabular}
    \caption{Mean of CPU time for 15000 iterations for the three problems. Implementation: Python language, Execution: PC with an Intel Core i7, 16 GB of RAM, Operating system: Windows. }
    \label{tab:time2}
\end{table}
\subsection{Analysis}
Table \ref{tab:time1} and \ref{tab:time2} show that the CPU time needed to solve the problems of each benchmark evolves linearly with respect to the number of circles to pack. Indeed, the more circles to position, the more forces to calculate within the virtual-force system of the CSO-VF algorithm. The required CPU time with the allocated numerical resources remains lower than 13.89 minutes for the first benchmark of problems (obtained for 55 circles) and lower than 98.7 minutes for the second benchmark of problems (obtained for 300 circles). It must be noted that the CPU time strongly depends on the implementation language used as well as the numerical resources allocated and is therefore hardly comparable with other CPU time from the literature. Furthermore, the CSO-VF algorithm is compatible with GPU programming that would considerably improve the computing time. However, from an industrial point of view, the effective computing times remain reasonable.

\newpage
 \bibliographystyle{elsarticle-num} 
 \bibliography{cas-refs}

\begin{thebibliography}{10}
\expandafter\ifx\csname url\endcsname\relax
  \def\url#1{\texttt{#1}}\fi
\expandafter\ifx\csname urlprefix\endcsname\relax\def\urlprefix{URL }\fi
\expandafter\ifx\csname href\endcsname\relax
  \def\href#1#2{#2} \def\path#1{#1}\fi

\bibitem{Grosso}
A.~Grosso, A.~Jamali, M.~Locatelli, F.~Shoen, Solving the problem of packing
  equal and unequal circles in a circular container, Journal of Global
  Optimization 47 (2010) 63--81.

\bibitem{Yuan}
Y.~Yuan, L.~Tole, F.~Ni, K.~He, Z.~Xiong, J.~Liu, Adaptive simulated annealing
  with greedy search for the circle bin packing problem, Computers \&
  Operations Research 144 (2022) 105826.

\bibitem{Stoyan}
Y.~Stoyan, G.~Yaskov, Packing equal circles into a circle with circular
  prohibited areas, International Journal of Computer Mathematics 89(10) (2012)
  1355--1369.

\bibitem{Bengtsson}
B.~Bengtsson, Packing rectangular pieces—a heuristic approach, The Computer
  Journal 25(3) (1982) 353--357.

\bibitem{Chen}
M.~Chen, C.~Wu, X.~Tang, X.~Peng, Z.~Zeng, S.~Liu, An efficient deterministic
  heuristic algorithm for the rectangular packing problem, Computers \&
  Industrial Engineering 137 (2019) 106097.

\bibitem{Bouzid}
M.~Bouzid, S.~Salhi, Packing rectangles into a fixed size circular container:
  Constructive and metaheuristic search approaches, European Journal of
  Operational Research 285(3) (2020) 865--883.

\bibitem{Daoden}
K.~Daoden, An adaptive no fit polygon (nfp) using modified sfla for the
  irregular shapes to solve the cutting and packing problem, International
  Journal of Advanced Science and Technology 29 (2020) 1046--1064.

\bibitem{Luo}
Q.~Luo, Y.~Rao, D.~Peng, Ga and gwo algorithm for the special bin packing
  problem encountered in field of aircraft arrangement, Applied Soft Computing
  114 (2022) 108060.

\bibitem{Guerriero}
F.~Guerriero, F.~Saccomanno, A hierarchical hyper-heuristic for the bin packing
  problem, Soft Computing (2022) 1--14.

\bibitem{Castillo}
I.~Castillo, F.~Kampas, J.~Pintér, Solving circle packing problems by global
  optimization: numerical results and industrial applications, European Journal
  of Operational Research 191(3) (2008) 786--802.

\bibitem{He-Tole}
K.~He, K.~Tole, F.~Ni, Y.~Yuan, L.~Liao, Adaptive large neighborhood search for
  solving the circle bin packing problem, Computers \& Operations Research 127
  (2021) 105140.

\bibitem{Huang}
W.~Huang, T.~Ye, Global optimization method for finding dense packings of equal
  circles in a circle, European Journal of Operational Research 210(3) (2011)
  474--481.

\bibitem{Galiev}
S.~Galiev, M.~Lisafina, Linear models for the approximate solution of the
  problem of packing equal circles into a given domain, European Journal of
  Operational Research 230(3) (2013) 505--514.

\bibitem{Lopez}
C.~López, J.~Beasley, A heuristic for the circle packing problem with a
  variety of containers, European Journal of Operational Research 214(3) (2011)
  512--525.

\bibitem{Birgin}
E.~Birgin, J.~Gentil, New and improved results for packing identical unitary
  radius circles within triangles, rectangles and strips, Computers \&
  Operations Research 37(7) (2010) 1318--1327.

\bibitem{He}
K.~He, D.~Mo, T.~Ye, W.~Huang, A coarse-to-fine quasi-physical optimization
  method for solving the circle packing problem with equilibrium constraints,
  Computers \& Industrial Engineering 66(4) (2013) 1049--1060.

\bibitem{Liu1}
J.~Liu, G.~Li, Basin filling algorithm for the circular packing problem with
  equilibrium behavioral constraints, Science China Information Sciences 53
  (2010) 885--895.

\bibitem{Sun}
Z.~Sun, H.~Teng, Optimal layout design of a satellite module, Engineering
  Optimization 35(5) (2003) 513--529.

\bibitem{Li2}
Z.~Li, Y.~Zeng, Y.~Wang, L.~Wang, B.~Song, A hybrid multi-mechanism
  optimization approach for the payload packing design of a satellite module,
  Applied Soft Computing 45 (2016) 11--26.

\bibitem{Gamot1}
J.~Gamot, M.~Balesdent, A.~Tremolet, R.~Wuilbercq, N.~Melab, E.-G. Talbi,
  Hidden-variables genetic algorithm for variable-size design space optimal
  layout problems with application to aerospace vehicles, Engineering
  Applications of Artificial Intelligence 121 (2023) 105941.

\bibitem{Mongeau}
M.~Mongeau, C.~Bes, Optimization of aircraft container loading, IEEE
  Transactions on Aerospace and Electronic Systems 39(1) (2003) 140--150.

\bibitem{Gajda}
M.~Gajda, A.~Trivella, R.~Mansini, D.~Pisinger, An optimization approach for a
  complex real-life container loading problem, Omega 107 (2022) 102559.

\bibitem{Caprace}
J.~Caprace, C.~Petcu, M.~Velarde, P.~Rigo, Optimization of shipyard space
  allocation and scheduling using a heuristic algorithm, Journal of Marine
  Science and Technology 18 (2013) 404--417.

\bibitem{Adickes}
M.~Adickes, R.~Billo, B.~Norman, S.~Banerjee, B.~Nnaji, J.~Rajgopal,
  Optimization of indoor wireless communication network layouts, Iie
  Transactions 34(9) (2002) 823--836.

\bibitem{Boonmee}
C.~Boonmee, M.~Arimura, T.~Asada, Facility location optimization model for
  emergency humanitarian logistics, International Journal of Disaster Risk
  Reduction 24 (2017) 485--498.

\bibitem{Herbert}
J.~Herbert-Acero, O.~Probst, P.~Réthoré, G.~Larsen, K.~Castillo-Villar, A
  review of methodological approaches for the design and optimization of wind
  farms, Energies 7(11) (2014) 6930--7016.

\bibitem{Xiao}
R.-B. Xiao, Y.-C. Xu, M.~Amos, Two hybrid compaction algorithms for the layout
  optimization problem, BioSystems 90(2) (2007) 560--567.

\bibitem{Xu}
Y.-C. Xu, R.-B. Xiao, M.~Amos, Particle swarm algorithm for weighted rectangle
  placement, In Third International Conference on Natural Computation (ICNC
  2007) 4 (2007) 728--732.

\bibitem{Costa}
A.~Costa, P.~Hansen, L.~Liberti, On the impact of symmetry-breaking constraints
  on spatial branch-and-bound for circle packing in a square, Discrete Applied
  Mathematics 161(1-2) (2013) 96--106.

\bibitem{Akeb}
H.~Akeb, M.~Hifi, R.~M'Hallah, A beam search algorithm for the circular packing
  problem, Computers \& Operations Research 36(5) (2009) 1513--1528.

\bibitem{He-Ye}
K.~He, H.~Ye, Z.~Wang, J.~Liu, An efficient quasi-physical quasi-human
  algorithm for packing equal circles in a circular container, Computers \&
  Operations Research 92 (2018) 26--36.

\bibitem{Zeng}
Z.~Zeng, X.~Yu, K.~He, W.~Huang, Z.~Fu, Iterated tabu search and variable
  neighborhood descent for packing unequal circles into a circular container,
  European Journal of Operational Research 250(2) (2016) 615--627.

\bibitem{Bello}
I.~Bello, H.~Pham, Q.~Le, M.~Norouzi, S.~Bengio, Neural combinatorial
  optimization with reinforcement learning, arXiv preprint arXiv:1611.09940
  (2016).

\bibitem{Kundu}
O.~Kundu, S.~Dutta, S.~Kumar, Deep-pack: A vision-based 2d online bin packing
  algorithm with deep reinforcement learning, In 2019 28th IEEE International
  Conference on Robot and Human Interactive Communication (RO-MAN) (2019) 1--7.

\bibitem{Zhao}
H.~Zhao, Q.~She, C.~Zhu, Y.~Yang, K.~Xu, Online 3d bin packing with constrained
  deep reinforcement learning, In Proceedings of the AAAI Conference on
  Artificial Intelligence 35(1) (2021) 741--749.

\bibitem{Hu}
H.~Hu, X.~Zhang, X.~Yan, L.~Wang, Y.~Xu, Solving a new 3d bin packing problem
  with deep reinforcement learning method, arXiv preprint arXiv:1708.05930
  (2017).

\bibitem{Tang}
F.~Tang, H.-F. Teng, A modified genetic algorithm and its application to layout
  optimization, Journal of Software 10(10) (1999) 1096--1102.

\bibitem{Teng}
H.-F. Teng, S.-L. Sin, D.-Q. Liu, Y.-Z. Li, Layout optimization for the objects
  located within a rotating vessel—a three-dimensional packing problem with
  behavioral constraints, Computers \& Operations Research 28(6) (2001)
  521--535.

\bibitem{Zhang}
B.~Zhang, H.-F. Teng, Y.-J. Shi, Layout optimization of satellite module using
  soft computing techniques, Applied Soft Computing 8(1) (2008) 507--521.

\bibitem{Liu2}
J.~Liu, G.~Li, D.~Chen, W.~Liu, Y.~Wang, Two-dimensional equilibrium constraint
  layout using simulated annealing, Computers \& Industrial Engineering 59(4)
  (2010) 530--536.

\bibitem{Liu3}
J.~Liu, G.~Li, H.~Geng, A new heuristic algorithm for the circular packing
  problem with equilibrium constraints, Science China Information Sciences 54
  (2011) 1572--1584.

\bibitem{Li}
Z.~Li, Z.~Tian, Y.~Xie, R.~Huang, J.~Tan, A knowledge-based heuristic particle
  swarm optimization approach with the adjustment strategy for the weighted
  circle packing problem, Computers \& Mathematics with Applications 66(10)
  (2013) 1758--1769.

\bibitem{Wang}
P.~Wang, S.~Huang, Z.-Q. Zhu, Swarm intelligence algorithms for circles packing
  problem with equilibrium constraints, In 2013 12th International Symposium on
  Distributed Computing and Applications to Business (2013) 55--60.

\bibitem{Moon}
I.~Moon, T.-V.-L. Nguyen, Container packing problem with balance constraints,
  OR Spectrum 36(4) (2014) 837--878.

\bibitem{Liu4}
J.~Liu, Y.~Jiang, G.~Li, Y.~Xue, Z.~Liu, Z.~Zhang, Heuristic-based energy
  landscape paving for the circular packing problem with performance
  constraints of equilibrium, Physica A: Statistical Mechanics and its
  Applications 431 (2015) 166--174.

\bibitem{Liu5}
J.~Liu, K.~Zhang, Y.~Yao, Y.~Xue, T.~Guan, A heuristic quasi-physical algorithm
  with coarse and fine adjustment for multi-objective weighted circles packing
  problem, Computers \& Industrial Engineering 101 (2016) 416--426.

\bibitem{Liu6}
J.~Liu, J.~Li, Z.~Lü, Y.~Xue, A quasi-human strategy-based improved basin
  filling algorithm for the orthogonal rectangular packing problem with mass
  balance constraint, Computers \& Industrial Engineering 107 (2017) 196--210.

\bibitem{Wang2}
Y.~Wang, Y.~Wang, J.~Sun, C.~Huang, X.~Zhang, A stimulus–response-based
  allocation method for the circle packing problem with equilibrium
  constraints, Physica A: Statistical Mechanics and its Applications 522 (2019)
  232--247.

\bibitem{Erbayrak}
S.~Erbayrak, V.~Özkır, U.-M. Yıldırım, Multi-objective 3d bin packing
  problem with load balance and product family concerns, Computers \&
  Industrial Engineering 159 (2021) 107518.

\bibitem{Romanova1}
T.~Romanova, Y.~Stoyan, A.~Pankratov, I.~Litvinchev, S.~Plankovskyy,
  Y.~Tsegelnyk, O.~Shypul, Sparsest balanced packing of irregular 3d objects in
  a cylindrical container, European Journal of Operational Research 291(1)
  (2021) 84--100.

\bibitem{Romanova2}
T.~Romanova, O.~Pankratov, I.~Litvinchev, P.~Stetsyuk, O.~Lykhovyd,
  J.~Marmolejo-Saucedo, P.~Vasant, Balanced circular packing problems with
  distance constraints, Computations 10(7) (2022) 113.

\bibitem{tarkesh}
H.~Tarkesh, A.~Atighehchian, A.~Nookabadi, Facility layout design using virtual
  multi-agent system, Journal of Intelligent Manufacturing 20 (2009) 347--357.

\bibitem{Ji}
K.~Ji, Q.~Zhang, Z.~Yuan, H.~Cheng, D.~Yu, A virtual force interaction scheme
  for multi-robot environment monitoring, Robotics and Autonomous Systems 149
  (2022) 103967.

\bibitem{Brambilla}
M.~Brambilla, E.~Ferrante, M.~Birattari, M.~Dorigo, Swarm robotics: a review
  from the swarm engineering perspective, Swarm Intelligence 7(1) (2013) 1--41.

\bibitem{Khatib}
O.~Khatib, J.~Burdick, Motion and force control of robot manipulators, In
  Proceedings. 1986 IEEE international conference on robotics and automation 3
  (1986) 1381--1386.

\bibitem{Tang99}
F.~Tang, H.~Teng, A modified genetic algorithm and its application to layout
  optimization, Journal of Software 10(10) (1999) 1096--1102.

\bibitem{Gamot2}
J.~Gamot, R.~Wuilbercq, M.~Balesdent, A.~Tremolet, N.~Melab, E.-G. Talbi,
  Component swarm optimization using virtual forces for solving layout
  problems, In Swarm Intelligence: 13th International Conference, ANTS 2022,
  Málaga, Spain, November 2–4, 2022, Proceedings Cham: Springer
  International Publishing. (2022) 292--299.

\bibitem{zhou}
C.~Zhou, L.~Gao, H.~Gao, Particle swarm optimization based algorithm for
  constrained layout optimization, Control Decision 20(1) (2005) 36--40.

\bibitem{shor}
N.~Shor, N.~Zhurbenko, A.~Likhovid, P.~Stetsyuk, P.i. algorithms of
  nondifferentiable optimization: Development and application, Cybernetics and
  System Analysis 39 (2003) 537--548.

\bibitem{stetsyuk}
P.~Stetsyuk, Shor’s r-algorithms: Theory and practice, In Optimization
  Methods and Applications: In Honor of Ivan V. Sergienko’s 80th Birthday;
  Butenko, S., Pardalos, P.M., Shylo, V. Springer: Berlin/Heidelberg, Germany
  (2017) 495--520.

\end{thebibliography}





% \end{thebibliography}
\end{document}